\documentclass[conference]{IEEEtran}
\IEEEoverridecommandlockouts
\usepackage{cite}
\usepackage{amsmath,amssymb,amsfonts}
\usepackage{algorithmic}
\usepackage{graphicx}
\usepackage{textcomp}
\usepackage{xcolor}

\usepackage{graphicx}
\usepackage{subcaption}  

\usepackage{amsmath,amssymb}
\usepackage{siunitx}

\usepackage{booktabs}
\usepackage{multirow}
\usepackage{array}
\usepackage{makecell}
\renewcommand{\arraystretch}{1.15}
\usepackage[hyphens]{url}
\usepackage{microtype}
\usepackage{array,tabularx,booktabs,multirow,makecell}
\usepackage[table]{xcolor}
\usepackage{microtype}
\newcolumntype{Y}{>{\raggedright\arraybackslash}X}
\newcolumntype{P}[1]{>{\raggedright\arraybackslash}p{#1}}

\definecolor{taskbin}{RGB}{222,237,255}   
\definecolor{task5c}{RGB}{255,235,205}    
\definecolor{taskmul}{RGB}{235,222,255}   
\definecolor{taskfl}{RGB}{218,247,220}    

\renewcommand{\arraystretch}{1.12}
\usepackage{array,tabularx,booktabs, multirow,makecell}
\usepackage{microtype}
\newcolumntype{L}{>{\raggedright\arraybackslash}X}  
\newcolumntype{C}{>{\centering\arraybackslash}p{1.25cm}} 

\usepackage{array,tabularx}
\newcolumntype{Y}{>{\raggedright\arraybackslash}X}   
\newcolumntype{P}[1]{>{\raggedright\arraybackslash}p{#1}} 

\usepackage{graphicx}
\usepackage{caption}
\usepackage{subcaption}

\usepackage{xcolor}

\def\BibTeX{{\rm B\kern-.05em{\sc i\kern-.025em b}\kern-.08em
    T\kern-.1667em\lower.7ex\hbox{E}\kern-.125emX}}
\begin{document}

\title{From Retinal Pixels to Patients: Evolution of Deep Learning Research in Diabetic Retinopathy Screening%
\thanks{This research has been funded by the Federal Ministry of Education and Research of Germany and the state of North-Rhine Westphalia as part of the Lamarr Institute for Machine Learning and Artificial Intelligence.}
}
\author{
\IEEEauthorblockN{
Muskaan Chopra\IEEEauthorrefmark{2}\IEEEauthorrefmark{4}, 
Lorenz Sparrenberg\IEEEauthorrefmark{2}\IEEEauthorrefmark{4}, Armin Berger\IEEEauthorrefmark{9}\IEEEauthorrefmark{2}\IEEEauthorrefmark{4},\\
Sarthak Khanna\IEEEauthorrefmark{2}, Jan H. Terheyden\IEEEauthorrefmark{6}, 
Rafet Sifa\IEEEauthorrefmark{9}\IEEEauthorrefmark{2}\IEEEauthorrefmark{4}
} \\
\IEEEauthorblockA{\IEEEauthorrefmark{9}Fraunhofer IAIS - Department of Media Engineering, Germany}
\IEEEauthorblockA{\IEEEauthorrefmark{2}University of Bonn - Department of Computer Science, Germany}
\IEEEauthorblockA{\IEEEauthorrefmark{6}University Hospital Bonn - Department of Ophthalmology, Germany \\}
\IEEEauthorblockA{\IEEEauthorrefmark{4}Lamarr Institute for Machine Learning and Artificial Intelligence, Germany}

}

\maketitle

\begin{abstract}
Diabetic Retinopathy (DR) remains a leading cause of preventable blindness, with early detection critical for reducing vision loss worldwide. Over the past decade, deep learning has transformed DR screening, progressing from early convolutional neural networks trained on private datasets to advanced pipelines addressing class imbalance, label scarcity, domain shift, and interpretability. This survey provides the first systematic synthesis of DR research spanning 2016-2025, consolidating results from 50+ studies and over 20 datasets. We critically examine methodological advances, including self- and semi-supervised learning, domain generalization, federated training, and hybrid neuro-symbolic models, alongside evaluation protocols, reporting standards, and reproducibility challenges. Benchmark tables contextualize performance across datasets, while discussion highlights open gaps in multi-center validation and clinical trust. By linking technical progress with translational barriers, this work outlines a practical agenda for reproducible, privacy-preserving, and clinically deployable DR AI. Beyond DR, many of the surveyed innovations extend broadly to medical imaging at scale.
\end{abstract}


\section{Introduction}
Diabetic Retinopathy (DR) is a leading cause of preventable blindness; early detection is critical to reduce vision loss. According to the International Diabetes Federation (IDF), approximately 537 million adults (aged 20-79) are currently living with diabetes, and this number is projected to rise to 643 million by 2030 \cite{IDFAtlas2021}. The World Health Organization and large epidemiological studies estimate that over one-third of people with diabetes will develop some form of DR during their lifetime \cite{Lee2015EpidemiologyDR}. Despite effective treatments such as laser photocoagulation and anti-VEGF therapy, clinical outcomes remain strongly dependent on early detection and diagnosis. Regular screening is therefore essential. However, health systems worldwide face significant challenges in meeting the escalating demand for retinal examinations. In recent years, advances in artificial intelligence and the increasing availability of large-scale retinal imaging datasets have created new opportunities to address these screening gaps at scale.

\subsection{Clinical context and grading}
Diagnosis of DR predominantly relies on retinal fundus photography, graded according to the International Clinical Diabetic Retinopathy (ICDR) Severity Scale, which includes five stages from No DR to Proliferative DR \cite{Wilkinson2003ICDR}. Many screening programs simplify this into a binary classification of referable versus non-referable DR to support operational workflows, though full grading remains important for research and clinical prognosis \cite{Abramoff2016IOVS, Lam2018DRPublic}. Recent reviews, such as Yang \textit{et al.} (2022) \cite{Yang2022Classification}, provide a broader historical perspective on DR classification systems, outlining how current ICDR/ETDRS scales capture clinical severity yet overlook neurodegenerative changes, and emerging imaging modalities. Their analysis highlights why ongoing refinement of grading standards is essential in parallel with advances in AI-based screening.

\begin{figure*}[!htbp]
\centering
\includegraphics[width=0.95\textwidth]{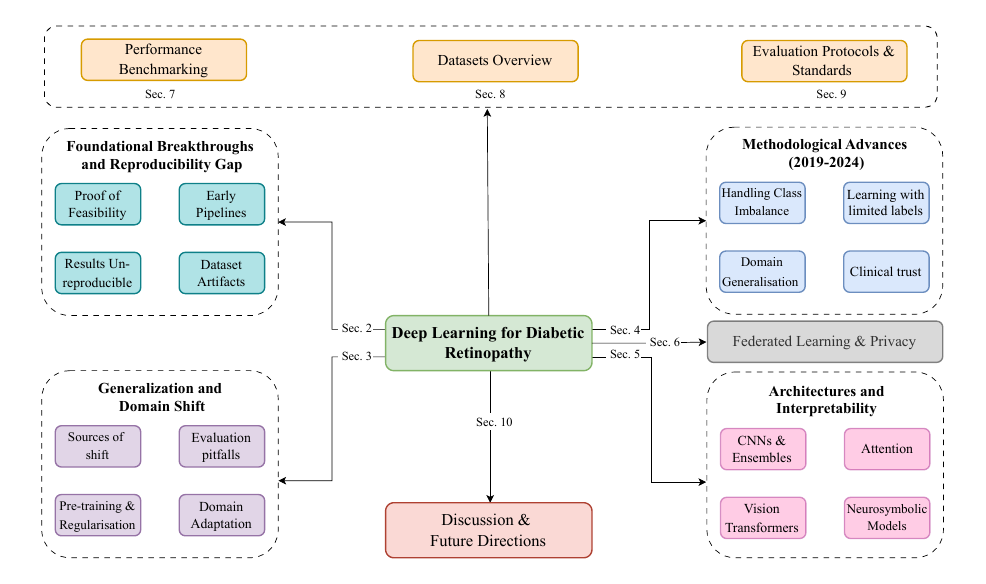}
\caption{Taxonomy and structure of this survey. 
The survey begins with foundational breakthroughs and the reproducibility gap (Section~\ref{sec:foundations}), 
addresses challenges of generalization and domain shift (Section~\ref{sec:domain_shift}), 
and methodological advances from 2019 - 2024 (Section~\ref{sec:methods}), 
then examines architectures and interpretability (Section~\ref{sec:arch_xai}), 
federated learning and privacy (Section~\ref{sec:federated}), 
datasets (Section~\ref{sec:datasets}), 
performance benchmarking (Section~\ref{sec:bench}), 
evaluation protocols (Section~\ref{sec:evaluation}), 
and finally discussion and future directions (Section~\ref{sec:discussion}).}
\label{fig:taxonomy}
\end{figure*}

\subsection{Promise of deep learning}
Deep learning (DL) rapidly emerged as a powerful approach for DR screening. Gulshan \textit{et al.}\ (2016) \cite{Gulshan2016JAMA} showed near-expert performance on large private datasets, followed by Krause \textit{et al.}\ (2018) \cite{Krause2018Ophthalmology} and Rakhlin \textit{et al.}\ (2018) \cite{Rakhlin2018ArXiv}, who extended DL to multi-task grading and related applications. These early successes positioned CNNs as a scalable tool for population-level screening. Yet, translation into practice required more than high accuracy; addressing data bottlenecks, integrating into screening programs, and obtaining regulatory approval remained critical hurdles \cite{GoogleBlog2016DR}.

\subsection{Challenges in early research}
Despite their promise, early models also revealed important limitations:

\begin{itemize}
    \item \textbf{Reproducibility crisis:} Many pioneering studies relied on proprietary datasets and unreleased code, complicating replication and fair benchmarking. This highlighted the urgent need for transparency, open-source implementations, and publicly accessible resources \cite{Voets2019Repro}.

    \item \textbf{Generalization gap:} Models that performed well in internal validation often showed degraded performance on external datasets due to demographic variability, device differences, and domain shifts. This exposed the necessity of diverse multi-center cohorts and standardized evaluation protocols to ensure reliable deployment \cite{Papadopoulos2021MIL}.

    \item \textbf{Task simplification:} Early works commonly reduced grading to a binary referable versus non-referable classification, neglecting the richer five-class clinical grading system. While operationally convenient, this limits prognostic insight and clinical adoption beyond initial screening \cite{Farag2020DenseNetCBAM, Taufiqurrahman2020MobileNetSVM}.

    \item \textbf{Black-box nature:} The lack of model interpretability hindered clinical trust and regulatory approval. Without lesion-level explanations or uncertainty quantification, clinicians were cautious about relying solely on automated outputs \cite{Sayres2019Ophthalmology}.
\end{itemize}

\subsection{Objectives and contributions of this survey}
The rapid advancement and diverse nature of research into deep learning (DL) methods for diabetic retinopathy (DR) screening have created an urgent need for a comprehensive and systematic synthesis of the field. This survey aims to fulfill this need by providing a critical overview of the evolution, current state, and future trajectory of DL-based approaches to DR screening. The contributions of this work are as follows:

\begin{enumerate}
\item \textbf{Clinical Trust and Reproducibility:} We perform a rigorous analysis of reproducibility challenges observed in early DL studies, highlighting recent progress toward open-source implementations, public dataset availability, and standardized evaluation practices that collectively advance clinical reliability.
\item \textbf{Benchmarking and Datasets:} This survey presents a consolidated benchmarking of representative DR screening methods published between 2016 and 2025 (Section~\ref{sec:bench}) and a curated summary of prominent retinal imaging datasets (Section~\ref{sec:datasets}). These resources facilitate contextual understanding of algorithmic performance relative to dataset characteristics and evaluation frameworks.
\item \textbf{Emerging Research Directions:} We discuss emerging methodological trends, such as self-supervised representation learning, federated and domain generalization techniques, and interpretable modeling strategies incorporating clinical expertise. The translational potential of these innovations for large-scale, real-world screening contexts is emphasized.
\end{enumerate}

Beyond diabetic retinopathy, many of the reviewed advances, such as self-supervised and federated learning, generalize to other medical imaging domains. Unlike earlier 2022–2023 surveys, this work integrates 2016–2025 studies under one taxonomy linking reproducibility, generalization, and clinical deployment.

The structure of this survey is depicted in Fig.~\ref{fig:taxonomy}, which maps the progression from foundational advances, through identification of key challenges, to subsequent innovations and current efforts aimed at clinical integration.

\subsection{Organization of the paper}
The remainder of this survey is organized as follows. Section~\ref{sec:foundations} introduces foundational CNN-based approaches and outlines the reproducibility crisis. Section~\ref{sec:domain_shift} discusses generalization challenges under domain shift, while Section~\ref{sec:methods} reviews methods addressing data scarcity, label noise, and class imbalance. Section~\ref{sec:arch_xai} surveys architectural innovations and interpretability techniques, and Section~\ref{sec:federated} covers federated and privacy-preserving training. Section~\ref{sec:datasets} and Section~\ref{sec:bench} consolidate datasets and benchmarking results. Finally, Section~\ref{sec:discussion} provides a critical discussion, and Section~\ref{sec:conclusion} concludes with future directions.

\section{Foundational Breakthroughs and the Reproducibility Gap}
\label{sec:foundations}

\begin{figure}[!htbp]
  \centering
  \begin{subfigure}[t]{0.3\columnwidth}
    \centering
    \includegraphics[width=\linewidth]{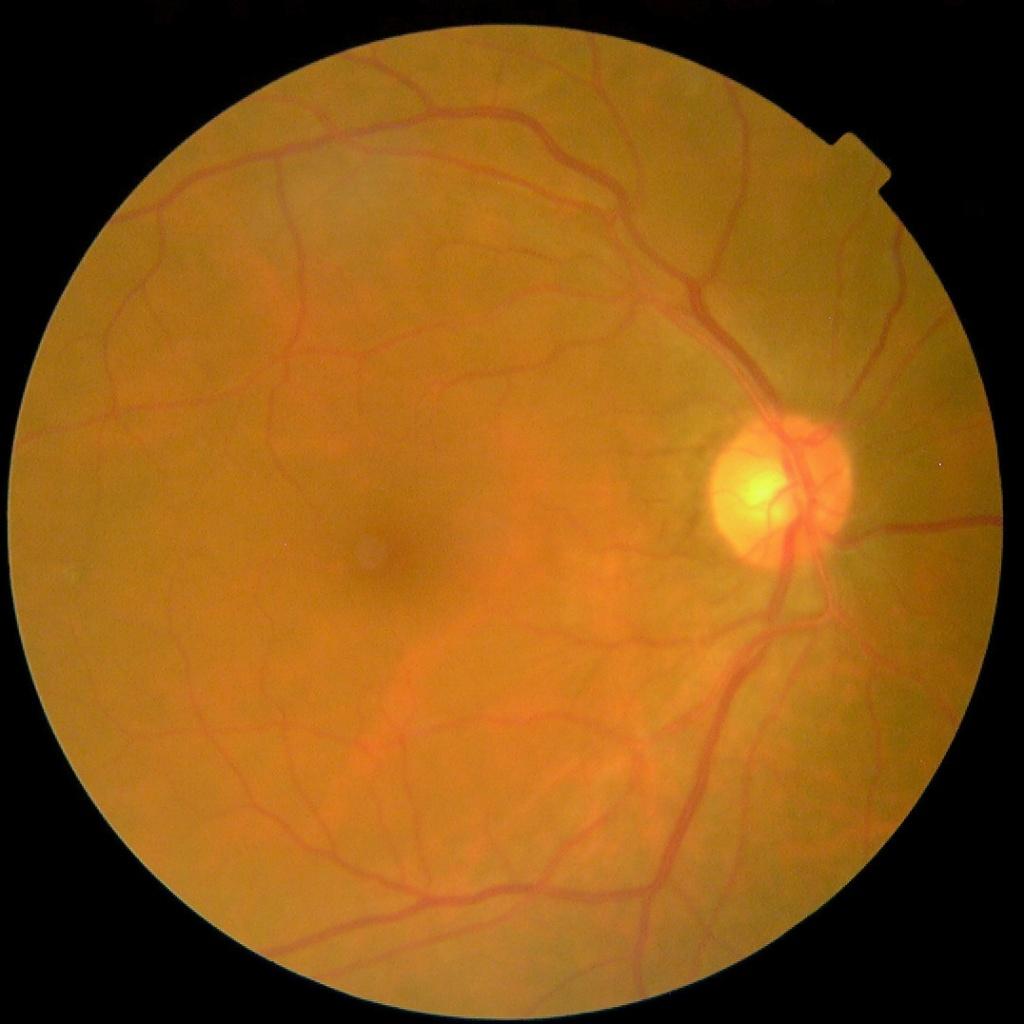}
    \caption{No DR}
    \label{fig:img1}
  \end{subfigure}
  \hspace{0.02\columnwidth} 
  \begin{subfigure}[t]{0.3\columnwidth}
    \centering
    \includegraphics[width=\linewidth]{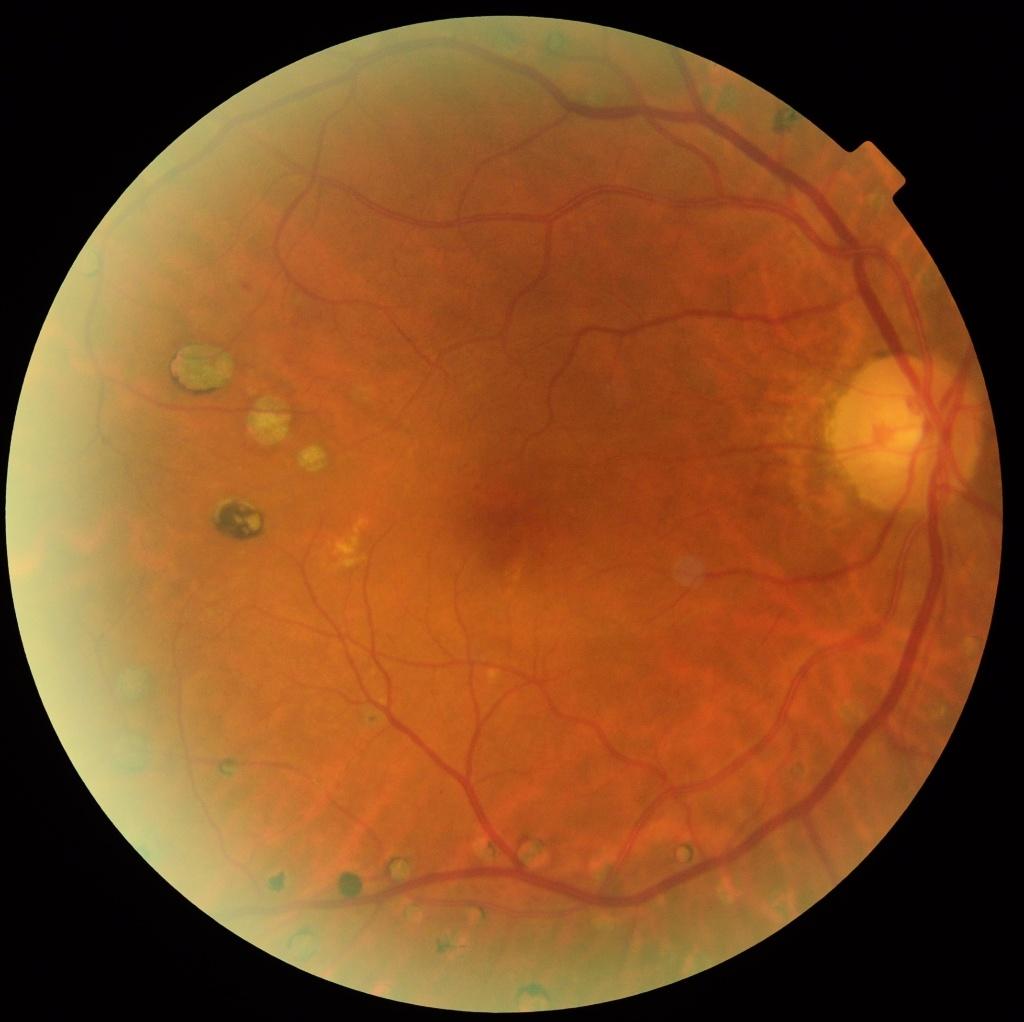}
    \caption{Severe DR}
    \label{fig:img2}
  \end{subfigure}
  \caption{Example fundus photographs \cite{Kaggle2015EyePACS} illustrating the spectrum of diabetic retinopathy.}

  \label{fig:two_images}
\end{figure}

\textbf{Proof of feasibility.}
The 2016 JAMA study by Gulshan \textit{et al.} \cite{Gulshan2016JAMA} demonstrated that convolutional neural networks (CNNs), trained on a very large internally curated fundus corpus, could detect referable diabetic retinopathy (rDR) at near-expert performance (AUC~$\approx$~0.99) on EyePACS and Messidor-2 datasets (refer Section~\ref{sec:datasets}; see Fig.~\ref{fig:two_images} for representative fundus examples). This landmark result catalyzed a wave of DL-for-DR research and paved the way for clinical systems. Around the same time, Abr{\`a}moff \textit{et al.} (2016) \cite{Abramoff2016IOVS} demonstrated high AUC ($\sim$0.98) on Messidor-2 using the proprietary IDx-DR system, highlighting the potential for regulatory-grade screening devices. Concurrent work investigated multi-task prediction (e.g., DR grade, diabetic macular edema, and image quality) and grader variability \cite{Krause2018Ophthalmology}, further normalizing DL as a component of screening programs.

\textbf{Task formulation and early pipelines.}
Most early systems framed screening as binary rDR prediction, a clinically aligned triage decision that is easier than fine-grained five-class grading. Kaggle-style pipelines (e.g., Lam \textit{et al.}, 2018; Rakhlin \textit{et al.}, 2018 \cite{Lam2018DRPublic, Rakhlin2018ArXiv}) experimented with contrast normalization, cropping to the retinal disk, and heavy ensembling (CNN+RNN hybrids). While these competition-driven strategies often reported strong validation accuracy, they rarely included external generalization tests, making it difficult to isolate the key factors contributing to model robustness.

\textbf{Why numbers were hard to reproduce.}
Independent re-implementations on public data reported substantially lower performance than originally claimed. Voets \textit{et al.} \cite{Voets2019Repro} showed that reproducing \cite{Gulshan2016JAMA} on EyePACS/Messidor-2 datasets with publicly available labels yielded AUCs in the 0.85-0.95 range, depending on label source and split policy. Three recurring factors were identified:
\begin{enumerate}
    \item \emph{Dataset opacity}: private curation steps and exclusion criteria were not fully described;
    \item \emph{Label inconsistency}: single-grader labels or differing adjudication rules between datasets introduced systematic shifts;
    \item \emph{Evaluation leakage}: per image rather than per-patient splits inflated metrics when both eyes from a patient appeared across train/test.
\end{enumerate}

\textbf{Clinical signal vs.\ dataset artifacts.}
High internal accuracy sometimes reflected sensitivity to acquisition factors (camera brand, illumination, field-of-view) rather than pathology. Strong quality control and preprocessing (e.g., Contrast Limited Adaptive Histogram Equalization (CLAHE)) improved internal validation yet induced artifacts that failed to translate across sites. Several studies reported sharp drops when re-tested on Messidor or DDR datasets after training on EyePACS/APTOS datasets (refer section \ref{sec:datasets}), highlighting \emph{domain shift} as the central obstacle to clinical portability \cite{Farag2020DenseNetCBAM, alwakid2023enhancing}. Domain shift remains a core methodological challenge across medical imaging.

\textbf{Takeaways for the community.}
The foundational period established (1) DL’s raw capability for fundus screening, and (2) the necessity of \emph{transparent} data curation, \emph{patient-level} evaluation, and \emph{external} testing for credible progress. Early interpretability attempts such as Sayres \textit{et al.} (2019) \cite{Sayres2019Ophthalmology} also revealed the limitations of post-hoc saliency maps, motivating the field’s later shift toward integrated attention mechanisms. These foundational insights have influenced not only DR research but also the broader development of trustworthy clinical AI methodologies.

\section{Generalization and Domain Shift}
\label{sec:domain_shift}

\textbf{Sources of shift.}
DR datasets differ along multiple axes: camera hardware (sensor, optics, resolution), capture protocol (mydriasis, field-of-view), demographics (age, ethnicity, comorbidities), prevalence of severe/proliferative diabetic retinopathy (PDR) cases, and label protocol (number of graders, adjudication). A model tuned to the EyePACS distribution can underperform when moved to Messidor-2 or DDR even with identical architectures and training schedules \cite{Kaggle2015EyePACS,Messidor2,DDR2019}. Beyond covariate shift, \emph{label shift} (different grading rubrics) and \emph{prior shift} (class frequency) further complicate transfer \cite{Wilkinson2003ICDR,Lee2015EpidemiologyDR}.

\textbf{Evaluation pitfalls.}
Per-image reporting inflates results when the two eyes of a patient are split across folds; class-imbalance can allow very high accuracy with poor PDR sensitivity; and strong pre/post-processing may bake in site-specific artifacts. We recommend (i) per-\emph{patient} splits and metrics, (ii) explicit thresholds with calibration (ECE/Brier), (iii) confidence intervals, and (iv) at least one \emph{external} test cohort \cite{Krause2018Ophthalmology,Sayres2019Ophthalmology,Voets2019Repro}.

\textbf{What helps in practice.}
Three families of techniques consistently improve portability:
\begin{enumerate}
    \item \textbf{Pretraining and regularization.} Large-scale self-supervised pretraining (contrastive or masked autoencoding) on diverse unlabeled fundus images learns features less tied to a single site and reduces label requirements \cite{Arrieta2023SPIE,Yang2024PLOSONE,Huang2021MICCAI,Alam2023Contrastive}. Favor simple, robust augmentations (including color jitter, mild blur, and crop/resize augmentations) over heavy image manipulation that can embed domain-specific artifacts; GAN-based augmentation should be validated with an external cohort \cite{Lim2020EMBC}.
    \item \textbf{Domain generalization (DG) and adaptation (DA).} Representation disentanglement and feature normalization (e.g., DECO) explicitly separate disease signal from domain cues, improving cross-dataset accuracy without target labels \cite{Xia2024MICCAI}. When unlabeled target data are available, consistency and pseudo-labeling stabilize adaptation. GDRNet (Che \textit{et al.}, 2023) \cite{Che2023GDRNet} and other DA pipelines \cite{Zhang2022DA,che2022learning} showed measurable cross-domain improvements.
    \item \textbf{Structure and priors.} Injecting retinal structure-vessels, optic disc/macula geometry, and lesion ontologies constrain the hypothesis space. Neuro-symbolic hybrids \cite{Urooj2025NeuroSymbolic} and lesion-aware attention discourage shortcuts (camera edges, illumination rings) and align evidence with clinical reasoning.
\end{enumerate}

DG/DA methods can still overfit to the collection of source domains; synthetic style-transfer may improve average accuracy while harming worst-case groups; and privacy constraints limit broad data pooling. These motivate federated learning (Sec.~\ref{sec:federated}), standardized reporting, and multi-center prospective validation.

\section{Methodological Advances (2019 - 2024)}
\label{sec:methods}

Following the recognition of core challenges, research from 2019 onward produced a rich set of methodological innovations. These approaches directly targeted issues of imbalance, data scarcity, domain shift, and clinical trust.

\subsection{Handling Class Imbalance}
Advanced stages of DR (Severe NPDR, PDR) are rare, making datasets highly skewed. Standard CNNs tend to collapse to majority classes, under-detecting sight-threatening disease. To mitigate this:
\begin{itemize}
    \item \textbf{Loss functions:} Class-balanced focal loss and cost-sensitive weighting improved minority recall without excessively harming majority accuracy.
    \item \textbf{Synthetic augmentation:} GAN-based synthetic image generation has been used to augment minority classes. Lim \textit{et al.} (2020) \cite{Lim2020EMBC} reported 3 - 5\% accuracy improvements on EyePACS, though concerns remain about unrealistic artifacts. Other augmentation pipelines such as ESRGAN super-resolution and CLAHE preprocessing \cite{alwakid2023enhancing} pushed performance on APTOS, but later work showed such gains often reflected artifacts rather than pathology \cite{Alwakid2022ESRGAN}.
    \item \textbf{Ensembles:} Qin \textit{et al.} (2022) \cite{qin2023classification} explored “deep forest” ensembles, combining multiple classifiers for robustness, but scalability is limited. Subsequent ensembles with EfficientNet backbones \cite{Arora2024SciRep} and self-adaptive stacking with attention \cite{Bodapati2023Stacking} achieved strong performance on Kaggle/EyePACS but introduced challenges for reproducibility due to architectural complexity.
\end{itemize}
Despite progress, true resolution requires larger annotated cohorts and careful per-patient balancing.

\subsection{Learning with Limited Labels}
Expert grading is costly; datasets like EyePACS include hundreds of thousands of unlabeled images. To leverage these:
\begin{itemize}
    \item \textbf{Semi-supervised learning (Semi-SL):} Teacher-student frameworks and pseudo-labeling, e.g., Duan \textit{et al.} (2022) \cite{Duan2022GACNN}, boosted performance on APTOS with 50\% fewer labels.
    \item \textbf{Self-supervised learning (SSL):} Contrastive methods such as lesion-aware pretraining \cite{Huang2021MICCAI,Alam2023Contrastive} yielded robust representations transferable to external cohorts. Arrieta \textit{et al.} (2023) \cite{Arrieta2023SPIE} showed EyePACS AUC = 0.94 using only 2\% labeled data.
    \item \textbf{Masked autoencoders (MAE):} Vision Transformers (ViT) pretraining with MAE scales effectively. Yang \textit{et al.} (2024) \cite{Yang2024PLOSONE} achieved AUC 0.98 with significantly fewer labels compared to ImageNet-pretrained CNNs. Related innovations such as equivariant refinement \cite{Fan2024ERCN} also emphasized label efficiency and robustness.
\end{itemize}

\subsection{Domain Generalization and Adaptation}
Generalization across datasets remains a core challenge. Approaches range from domain adaptation methods (e.g., pseudo-labeling, adversarial training, and curriculum strategies \cite{Zhang2022DA,che2022learning}) to domain generalization techniques such as DECO \cite{Xia2024MICCAI} and GDRNet \cite{Che2023GDRNet}, which disentangle acquisition factors or improve performance on unseen datasets. More recently, neuro-symbolic hybrids \cite{Urooj2025NeuroSymbolic} integrate ViTs with lesion ontologies, achieving notable gains in cross-domain robustness.

\subsection{Building Clinical Trust}
Beyond accuracy, clinicians demand interpretability and safety:
\begin{itemize}
    \item \textbf{Explainable AI:} Post-hoc heatmaps (Grad-CAM, IG) provided lesion-localization cues \cite{Sayres2019Ophthalmology}, but can be misleading. Efforts such as ULBPEZ feature encodings \cite{Shakeel2022ULBPEZ} sought interpretable handcrafted features with mixed success.
    \item \textbf{Integrated attention:} IDANet (Bhati \textit{et al.}, 2024) \cite{Bhati2024AIM} embedded dual attention to highlight pathologic regions, improving transparency. More recent dual-attention CNNs \cite{Hannan2025DualAttention} extended this idea to handle class imbalance explicitly.
    \item \textbf{Uncertainty estimation:} Bayesian CNNs and MC dropout quantified model confidence \cite{akram2025uncertainty}, supporting safe referral thresholds. Other works incorporated calibration metrics explicitly during training \cite{Papadopoulos2021MIL}.
\end{itemize}

 Methodological advances shifted the field from proof-of-concept CNNs to pipelines capable of learning from scarce labels, transferring across domains, and providing clinically relevant explanations. However, robust multi-center validation remains the exception rather than the norm.

\section{Architectures and Interpretability}
\label{sec:arch_xai}

\subsection{From CNN Backbones to Ensembles}
Early DR pipelines relied on CNNs such as Inception-v3 and ResNet-50, pretrained on ImageNet and fine-tuned for fundus images. Simple preprocessing (e.g., optic disc cropping, histogram equalization) helped establish strong baselines. To boost accuracy, ensembles became common: Arora \textit{et al.} (2024) \cite{Arora2024SciRep} showed EfficientNet ensembles achieved $\sim$86.5\% balanced accuracy on EyePACS with improved calibration. However, ensembles raise computational costs, limiting scalability and reproducibility. Fusion-based transfer learning \cite{Aftab2025Fusion} and adaptive stacking with attention \cite{Bodapati2023Stacking} show promise but share these concerns.

\subsection{Attention Mechanisms}
Attention modules emphasize pathological regions while reducing background noise. Dual-attention CNNs (Bhati \textit{et al.}, 2024; Hannan \textit{et al.}, 2025) \cite{Bhati2024AIM,Hannan2025DualAttention} and refinement-based approaches \cite{wang2025diabetic} improve lesion localization, though reliance on small or single-center datasets limits generalizability.

\subsection{Vision Transformers (ViTs)}
Inspired by NLP, ViTs use self-attention over image patches to capture global context. Huang \textit{et al.} (2021) \cite{Huang2021MICCAI} showed lesion-based contrastive pretraining improved EyePACS generalization, while Yang \textit{et al.} (2024) \cite{Yang2024PLOSONE} achieved state-of-the-art AUC (0.98) with masked autoencoder pretraining using fewer labels. 
These models outperform CNNs mainly through global-context attention and large-scale pretraining \cite{Fan2024ERCN} , which improve cross-dataset generalization, though CNNs still excel on smaller sets due to stronger inductive bias.

\subsection{Neuro-Symbolic and Hybrid Models}
Newer models combine neural networks with symbolic priors. Urooj \textit{et al.} (2025) \cite{Urooj2025NeuroSymbolic} integrated ViTs with vascular and lesion ontologies, boosting cross-domain performance by more than 5\%. Zhou \textit{et al.} (2025) \cite{Zhou2025GPMKLE} proposed GPMKLE-Net with self-paced multi-scale learning, reflecting a trend toward clinician-aligned, generalizable systems.

\subsection{Interpretability and Clinical Trust}
Explainability remains a barrier. Post-hoc methods (e.g., Grad-CAM, integrated gradients) can mislead by highlighting artifacts \cite{Sayres2019Ophthalmology}. Integrating attention directly into models, as in IDANet, yields more clinically consistent explanations. Uncertainty estimation (Bayesian CNNs, MC dropout) \cite{akram2025uncertainty} supports safe referral, while lightweight CNNs (Gayathri \textit{et al.}, 2020) \cite{gayathri2020lightweight} show potential for resource-limited settings but must balance interpretability, efficiency, and accuracy.

Architectural progress in DR AI has moved from conventional CNNs to attention-based, transformer-based, and hybrid neuro-symbolic models. The field is advancing from reliance on post-hoc interpretability toward integrated lesion-aware and uncertainty-driven designs. Nonetheless, multicenter validation and practical deployment require continued attention to the trade-offs between accuracy, efficiency, scalability, and clinical trust.

\section{Federated and Privacy-Preserving Learning}
\label{sec:federated}

\subsection{Why Federated Learning?}
Access to diverse, multi-center retinal datasets is crucial for developing generalizable models in diabetic retinopathy (DR) screening. However, privacy frameworks such as HIPAA and GDPR, as well as institutional policies, severely restrict the transfer of raw patient data across clinical sites. Centralized data pooling is thus often infeasible due to concerns around patient confidentiality, ethics, and logistical barriers. Federated Learning (FL) addresses this challenge by enabling institutions to collaboratively train models without sharing raw images: models are trained locally on institutional data, and only model updates are aggregated on a central server. This decentralized approach enables collaborative learning while maintaining compliance with privacy regulations and data governance requirements.

\subsection{Simulation Studies}
Initial DR-focused FL studies generally simulated multi-center scenarios using public datasets. Matta \textit{et al.} (2023) \cite{Matta2023SciRep} trained federated CNN ensembles on EyePACS, Messidor, and IDRiD, demonstrating only minor accuracy reductions compared to centralized setups. Mohan Raj \textit{et al.} (2024) \cite{MohanRaj2024FL} evaluated EfficientNet-based FL across DDR, EyePACS, and APTOS, reporting robust generalization ($\sim$93.2\% accuracy) and resilience to lower-quality images. Additional ensemble-based models, including Wong \textit{et al.} (2023) \cite{Wong2023ECOC} and Aftab \textit{et al.} (2025) \cite{Aftab2025Fusion}, showcase the potential to adapt fusion and stacking methods for federated or distributed learning, especially for multi-dataset and cross-domain training. Recent innovations such as FedGAN \cite{kamran2025fedgan} enable privacy-preserving generation of synthetic retinal images, further mitigating data scarcity and regulatory barriers.

\subsection{Challenges and Limitations}
Despite these advances, several significant challenges persist:
\begin{itemize}
\item \textbf{Statistical heterogeneity:} Non-IID distributions across institutions (variations in class prevalence, sensors, protocols) can weaken model convergence and performance.
\item \textbf{Communication overhead:} Training large models (such as ViTs or ensembles) exacerbates network bandwidth usage and synchronization delays relative to centralized learning.
\item \textbf{Security risks:} Even model updates may leak sensitive information unless augmented by differential privacy, homomorphic encryption, or secure aggregation protocols.
\item \textbf{Evaluation gaps:} Most published FL studies rely on simulations with public datasets, rather than true prospective clinical deployment and validation.
\end{itemize}

\subsection{Outlook}
Key priorities for future research in DR-FL include:
\begin{enumerate}
\item \textbf{Personalization:} Incorporating local adaptation layers to adjust for site-specific biases and data distributions.
\item \textbf{Efficiency:} Leveraging lightweight models (MobileViT, quantized CNNs, or efficient CNNs \cite{gayathri2020lightweight}) to minimize communication and computation costs without sacrificing accuracy.
\item \textbf{Security guarantees:} Integrating privacy-preserving cryptography such as differential privacy or secure aggregation to further protect sensitive information.
\item \textbf{Prospective validation:} Designing and reporting real-world FL deployments that bridge institutions and geographies, assessing robustness and clinical relevance in true clinical workflows.
\end{enumerate}

Federated learning represents a pivotal advance toward privacy-preserving, multi-center AI in DR screening. While current studies demonstrate feasibility, real-world deployment remains limited; regulatory, communication, and heterogeneity barriers must still be solved before FL achieves sustained multi-hospital validation.

\section{Datasets Overview}
\label{sec:datasets}

Table~\ref{tab:datasets} summarizes the most commonly used datasets for DR research, spanning public competitions, curated clinical cohorts, and private collections. In the following subsections, we group them into public competition benchmarks, clinical datasets, and private cohorts to highlight their respective roles and limitations.

\subsection{Public Competition Datasets}
The Kaggle EyePACS competition remains the most widely used benchmark. Over 30 of the papers we reviewed report results on EyePACS, though often with differing splits, label protocols, or data cleaning pipelines (e.g., Lam et al. 2018; Rakhlin et al. 2018; Yuan et al. 2020). These inconsistencies complicate fair comparison. APTOS 2019 \cite{APTOS2019}, with 3,662 labeled images, is another widely used benchmark, appearing in at least 15 surveyed works (Farag et al., 2022; Duan et al. 2022; Alwakid et al. 2023; Hannan et al. 2025). However, its small size and imbalance have led to extreme overfitting, with many works reporting suspiciously high accuracies (more than 97\%) that fail to generalize.

\begin{table*}[!htbp]
\scriptsize
\centering
\caption{Summary of commonly used diabetic retinopathy (DR) datasets. References correspond to dataset descriptions or first major use.}
\label{tab:datasets}
\begin{tabular}{p{2.2cm} p{0.9cm} p{2.2cm} p{2.2cm} p{1.6cm} p{6.2cm}}
\toprule
\textbf{Dataset} & \textbf{Year} & \textbf{Size} & \textbf{Labels / Task} & \textbf{Access} & \textbf{Notes / Citations} \\
\midrule
EyePACS (Kaggle) & 2015 & $\sim$88k images & 5-class; rDR & Public (Kaggle) & Most widely used; noisy labels, inconsistent grading \cite{Kaggle2015EyePACS}. \\
Messidor-1 & 2010 & 1,200 images & 4-level DR, rDR & Public & Older cameras; often merged with Messidor-2 \cite{Messidor2}. \\
Messidor-2 & 2014 & 1,748 images & 5-class; rDR & Public & Common external test set; used in \cite{Gulshan2016JAMA,Voets2019Repro}. \\
APTOS 2019 & 2019 & 3,662 images & 5-class & Public (Kaggle) & Severe imbalance; many works overfit ($>$97\% acc.) \cite{APTOS2019}. \\
DDR (China) & 2019 & 13,673 images & 5-class + lesions & Public & Diverse; common external validation \cite{DDR2019}. \\
IDRiD (India) & 2018 & 516 images & 5-class + lesion masks & Public & Gold standard for lesion segmentation + grading \cite{IDRiD2018DATA}. \\
FGADR & 2020 & 2,842 images & 5-class + lesions & Public & Rich annotations, modest scale \cite{FGADR2020SciData}. \\
ODIR  & 2019 & 5,000 images & Multi-disease incl.\ DR & Public & Mixed conditions; domain diversity \cite{ODIR2019}. \\
DIARETDB0/1 & 2006/07 & 130 + 89 images & Lesion segmentation & Public & Early DR test sets for algorithm development. \\
DRIVE & 2004 & 40 images & Vessel segmentation & Public & Small but widely used for retinal vessel analysis. \\
CHASE-DB1 & 2010 & 28 images & Vessel segmentation & Public & Dual-annotated vessel dataset. \\
STARE & 2000 & 400 images & Vessel/lesion segmentation & Public & Classic dataset for vessel detection/DR lesions. \\
HRF & 2013 & 45 images & Glaucoma, DR, DME segmentation & Public & High-resolution, mixed-pathology fundus images. \\
UIC Clinic & 2023 & 2500 images & Binary rDR & Private & Used for external validation in SSL generalization \cite{Alam2023Contrastive}\cite{uic}. \\
mBRSET (Brazil) & 2024 & 5164 images & 5-class DR & Public (PhysioNet) & Collected via low-cost mobile fundus camera; new public dataset.\cite{mbrset} \\
TJDR & 2023 & 561 images & 5-class + pixel-level masks & Public & Recent joint annotated dataset for classification/segmentation. \cite{mao2023tjdr} \\
Private Google & 2016 & 128k images & 5-class; rDR & Private & Used in \cite{Gulshan2016JAMA}; unreleased. \\
\bottomrule
\end{tabular}
\end{table*}

\subsection{Clinical Datasets}
Messidor-2 \cite{Messidor2} is the most common external validation set. Dozens of studies (e.g., Gulshan et al. 2016; Voets et al. 2019; Cheung et al. 2021; Che et al. 2023) use Messidor-2 as a hold-out test, often revealing substantial performance drops compared to Kaggle-split numbers. DDR (Chinese, 13,673 images) \cite{DDR2019} has become another external benchmark, featured in Akram et al. 2025; Bhati et al. 2024; Alwakid et al. 2023. IDRiD \cite{IDRiD2018DATA} and FGADR \cite{FGADR2020SciData} provide detailed lesion annotations, enabling hybrid classification-segmentation models (Li et al. 2019; Zhang et al. 2022). ODIR \cite{ODIR2019} further broadens the scope by including multi-disease fundus images. Other corpora like DIARETDB0/1, DRIVE, CHASE, STARE, and HRF remain useful for lesion/vessel segmentation tasks and early CNN baselines \cite{Bhimavarapu2022Activation}.

\subsection{Private Clinical Cohorts}
Several influential papers report training on massive private datasets (Google 2016 \cite{Gulshan2016JAMA}; Krause 2018 \cite{Krause2018Ophthalmology}). While these datasets enable large-scale training, their inaccessibility exacerbates the reproducibility crisis. Replication studies (Voets 2019 \cite{Voets2019Repro}; Papadopoulos 2021 \cite{Papadopoulos2021MIL}) show that results do not always carry over to public sets. Recent federated learning efforts (Matta 2023 \cite{Matta2023SciRep}; Mohan Raj 2024 \cite{MohanRaj2024FL}) highlight strategies to leverage local private cohorts without centralizing data.

In total, we identified over 20 distinct datasets across 50+ studies. Table~\ref{tab:datasets} consolidates key properties (size, labels, access). The lack of standardized splits and per-patient protocols remains a barrier: many reported accuracies are not directly comparable, underlining the need for community benchmarks and open science practices.

\section{Performance Benchmarking}
\label{sec:bench}

\subsection{Reported Metrics Across Studies}
Over the past decade, reported performance on DR benchmarks has ranged from AUC $>$0.99 in early private datasets \cite{Gulshan2016JAMA} to more modest accuracies ($\sim$85--90\%) on public sets with rigorous splits \cite{Voets2019Repro}. Table~\ref{tab:metrics} consolidates representative results (2016--2025) and lists a separate line per dataset to avoid mixing tasks or cohorts. Metrics are reproduced as reported in each paper. A dash (``--'') indicates the paper did not report that metric for the given dataset and task.

\subsection{Discussion of Reported Results}
Several trends are apparent:
\begin{itemize}
    \item \textbf{Inflated early results:} AUCs above 0.99 (e.g., Gulshan et al. 2016) were rarely reproduced on public test sets \cite{Voets2019Repro}.
    \item \textbf{Public competitions:} On APTOS 2019, dozens of papers report accuracies $>$90\% \cite{Duan2022GACNN,qin2023classification,Mohanty2023DLArch}, but generalization to Messidor/DDR is poor.
    \item \textbf{Label efficiency:} SSL and MAE pretraining (Arrieta 2023; Yang 2024; Fan 2024) achieved competitive or superior performance with fewer labels.
    \item \textbf{Trust-oriented models:} Attention-based (IDANet \cite{Bhati2024AIM}, dual-attention CNNs \cite{Hannan2025DualAttention}) and Bayesian approaches \cite{akram2025uncertainty} emphasize interpretability and safety.
    \item \textbf{Cross-domain generalization:} DECO \cite{Xia2024MICCAI}, GDRNet \cite{Che2023GDRNet}, and neuro-symbolic hybrids \cite{Urooj2025NeuroSymbolic} show promise in mitigating domain shift.
    \item \textbf{Deployment focus:} Lightweight CNNs \cite{gayathri2020lightweight} and FL pipelines \cite{Matta2023SciRep,MohanRaj2024FL} highlight practical trade-offs for clinical use.
\end{itemize}

\begin{table*}[!htbp]
\centering
\caption{Performance for DR screening (2016-2025). Pub = Public, Pvt = Private. Tasks: Bin = binary referable DR detection; 5c = 5-class DR grading; Multi = multi-task (e.g., DR+DME); FL = federated learning simulation. Acc. = Accuracy, Sens. = Sensitivity, Spec. = Specificity}
\label{tab:metrics}
\setlength{\tabcolsep}{8pt}
\renewcommand{\arraystretch}{1.15}
\begin{tabularx}{\textwidth}{l l c c c c c c}
\toprule
\textbf{Paper} & \textbf{Dataset (Pub/Pvt)} & \textbf{Task} & \textbf{AUC} & \textbf{Acc.} & \textbf{Kappa} & \textbf{Sens.} & \textbf{Spec.} \\
\midrule
\multicolumn{8}{l}{\textbf{2016--2018 (Foundational)}} \\
Gulshan et al.\ (2016) \cite{Gulshan2016JAMA}      & EyePACS (Pub)        & Bin   & 0.99 & -    & -            & 0.90      & 0.98 \\
Gulshan et al.\ (2016) \cite{Gulshan2016JAMA}      & Messidor-2 (Pub)     & Bin   & 0.99 & -    & -            & 0.87      & 0.98 \\
Abr{\`a}moff et al.\ (2016) \cite{Abramoff2016IOVS} & Messidor-2 (Pub)     & Bin   & 0.98 & -    & -            & 0.96 & 0.87 \\
Krause et al.\ (2018) \cite{Krause2018Ophthalmology}& Moorfields (Pvt)     & Multi & 0.98    & - & 0.84      & 0.97      & 0.92 \\
Lam et al.\ (2018) \cite{Lam2018DRPublic}          & EyePACS (Pub)        & Bin   & -    & 0.74 & -            & -      & - \\
Lam et al.\ (2018) \cite{Lam2018DRPublic}          & Messidor (Pub)       & 5c    & -    & 0.57 & -            & -      & - \\
Rakhlin et al.\ (2018) \cite{Rakhlin2018ArXiv}     & EyePACS (Pub)        & Bin    & 0.92    & -    & -  & 0.92  & 0.92 \\
Rakhlin et al.\ (2018) \cite{Rakhlin2018ArXiv}     & Messidor-2 (Pub)        & Bin    & 0.97    & -    & -  & 0.99  & 0.92 \\

\midrule
\multicolumn{8}{l}{\textbf{2019--2021 (Reproducibility \& Early SSL)}} \\
Voets et al.\ (2019) \cite{Voets2019Repro}         & EyePACS (Pub)        & Bin   & 0.95    & -    & -            & 0.90      & 0.83 \\
Voets et al.\ (2019) \cite{Voets2019Repro}         & Messidor-2 (Pub)     & Bin   & 0.85    & -    & -            & 0.81      & 0.68 \\
Sayres et al.\ (2019) \cite{Sayres2019Ophthalmology}& EyePACS/Messidor (Pub)& 5c & -    & -    & -            & 0.91 & 0.94 \\
Taufiqurrahman et al.\ (2020) \cite{Taufiqurrahman2020MobileNetSVM} & APTOS 2019 (Pub) & 5c & - & 0.85 & 0.92 & - & - \\
Lim et al.\ (2020) \cite{Lim2020EMBC}              & EyePACS (Pub)        & 5c    & 0.98    & -    & -            & -      & - \\
Saxena et al.\ (2020) \cite{saxena2020improved}    & Messidor (Pub) & Bin & 0.95 & - & - & 0.88 & 0.90 \\
Saxena et al.\ (2020) \cite{saxena2020improved}    & Messidor-2 (Pub) & Bin & 0.92 & - & - & 0.81 & 0.86 \\
Saxena et al.\ (2020) \cite{saxena2020improved}    & EyePACS (Pub) & Bin & 0.92 & - & - & 0.84 & 0.89 \\
Gayathri et al.\ (2020) \cite{gayathri2020lightweight} & EyePACS (Pub)    & Bin   & -    & 0.99 & -            & 1.00      & 1.00 \\
Gayathri et al.\ (2020) \cite{gayathri2020lightweight} & Messidor (Pub)    & Bin   & -    & 0.99 & -            & 0.99      & 0.99 \\
Gayathri et al.\ (2020) \cite{gayathri2020lightweight} & IDRiD (Pub) & Bin    & -    & 0.99 & -            & 0.99      & 0.99 \\
Huang et al.\ (2021) \cite{Huang2021MICCAI}        & EyePACS (Pub)        & 5c (SSL) & - & - & 0.83 & - & - \\
Papadopoulos et al.\ (2021) \cite{Papadopoulos2021MIL} & IDRiD & Bin & 0.86 & - & - & - & - \\
\midrule
\multicolumn{8}{l}{\textbf{2022-2023 (Imbalance, SSL, Domain Generalization)}} \\
Farag et al.\ (2022) \cite{Farag2020DenseNetCBAM}  & APTOS 2019 (Pub)     & 5c    & -    & 0.82 & 0.88            & -      & - \\
Farag et al.\ (2022) \cite{Farag2020DenseNetCBAM}  & APTOS 2019 (Pub)     & Bin    & -    & 0.97 & 0.94            & 0.97      & 0.98 \\
Duan et al.\ (2022) \cite{Duan2022GACNN}           & APTOS 2019 (Pub)     & 5c (Semi-SL) & - & 0.93 & 0.91 & 0.93 & - \\
Che et al.\ (2022) \cite{che2022learning}          & Messidor (Pub)       & 5c    & 0.86    & 0.70 & -           & -      & - \\
Che et al.\ (2022) \cite{che2022learning}          & IDRiD (Pub)          & 5c    & 0.84    & 0.59 & -           & -      & - \\
Berbar (2022) \cite{Shakeel2022ULBPEZ}             & EyePACS (Pub)        & Bin    & 0.97    & 0.97 & -          & 1.00  & 1.00 \\
Berbar (2022) \cite{Shakeel2022ULBPEZ}             & Messidor-2 (Pub)     & 3c   & -    & 0.98 & -          & 1.00  & 1.00 \\
Berbar (2022) \cite{Shakeel2022ULBPEZ}             & EyePACS (Pub)        & 3c    & -    & 0.96 & -          & 1.00  & 0.96 \\
Qin et al.\ (2023) \cite{qin2023classification}    & EyePACS (Pub)        & 5c    & -    & 0.74 & -            & 0.74      & - \\
Arrieta et al.\ (2023) \cite{Arrieta2023SPIE}      & EyePACS (Pub)        & Bin   & 0.94 & -    & -            & -      & - \\
Arrieta et al.\ (2023) \cite{Arrieta2023SPIE}      & Messidor-2 (Pub)     & Bin   & 0.89 & -    & -            & -      & - \\
Alwakid et al.\ (2023) \cite{alwakid2023enhancing} & APTOS 2019 (Pub)     & 5c    & -    & 0.98 & -           & 0.98      & - \\
Alam et al.\ (2023) \cite{Alam2023Contrastive}     & UIC private (Pvt, external) & Bin & 0.91 & - & - & - & - \\
Mohanty et al.\ (2023) \cite{Mohanty2023DLArch}    & APTOS 2019 (Pub)     & 5c    & -    & 0.97 & -            & -      & - \\
Wong et al.\ (2023) \cite{Wong2023ECOC}            & APTOS 2019 (Pub) & 5c & - & 0.82 & - & - & - \\
Wong et al.\ (2023) \cite{Wong2023ECOC}            & EyePACS+Messidor (Pub) & 3c & - & 0.75 & - & - & - \\
Bodapati \& Balaji (2023) \cite{Bodapati2023Stacking} & APTOS 2019 (Pub) & 5c & 0.96 & 0.86 & 0.89 & 0.72 & - \\
Matta et al.\ (2023) \cite{Matta2023SciRep}        & OPHDIAT & 5c (FL) & 0.95 & - & - & - & - \\
\midrule
\multicolumn{8}{l}{\textbf{2024-2025 (Transformers, Federated, Hybrids)}} \\
Arora et al.\ (2024) \cite{Arora2024SciRep}        & EyePACS (Pub)        & 5c    & -    & 0.86 & -           & -      & - \\
Shakibania et al.\ (2024) \cite{shakibania2024dual} & APTOS 2019 (Pub)    & 5c    & 0.89    & 0.83 & 0.89           & 0.83 & 0.94 \\
Shakibania et al.\ (2024) \cite{shakibania2024dual} & APTOS 2019 (Pub)    & Bin    & 0.97    & 0.97 & 0.95           & 0.98 & 0.97 \\
Bhati et al.\ (2024) \cite{Bhati2024AIM}           & DDR (Pub)    & 5c    & 0.86 & 0.85    & -            & 0.82      & - \\
Bhati et al.\ (2024) \cite{Bhati2024AIM}           & EyePACS (Pub)     & 5c    & 0.96 & 0.93    & -            & 0.92      & - \\
Bhati et al.\ (2024) \cite{Bhati2024AIM}           & IDRiD (Pub)     & 5c    & 0.93 & 0.90    & -            & 0.84      & - \\
Yang et al.\ (2024) \cite{Yang2024PLOSONE}         & APTOS 2019 (Pub)        & Bin   & 0.98 & 0.93 & -          & 0.96 & 0.95\\
Xia et al.\ (2024) \cite{Xia2024MICCAI}            & FGADR (Pub) & 5c & 0.86 & 0.57 & - & - & - \\
Fan et al.\ (2024) \cite{Fan2024ERCN}              & EyePACS (Pub) & 5c (SSL) & 0.93 & 0.87 & 0.85 & - & - \\
Mohan Raj et al.\ (2024) \cite{MohanRaj2024FL}     & EyePACS (Pub) & 5c (FL) & - & 0.90 & - & - & - \\
Urooj et al.\ (2025) \cite{Urooj2025NeuroSymbolic} & EyePACS / APTOS / Messidor (Pub) & 5c (Hybrid) & - & 0.63 & - & - & - \\
Wang et al.\ (2025) \cite{wang2025diabetic}        & APTOS 2019 (Pub)     & 5c    & 0.89    & - & 0.82           & 0.80      & 0.81 \\
Zhou et al.\ (2025) \cite{Zhou2025GPMKLE}          & APTOS + Messidor (Pub) & 5c  & 0.99    & 0.94    & -           & -      & - \\
Aftab et al.\ (2025) \cite{Aftab2025Fusion}        & APTOS + IDRiD + Messidor-2 (Pub) & Bin / 5c & - & 0.96  & - & - & - \\
Hannan et al.\ (2025) \cite{Hannan2025DualAttention}& APTOS 2019 (Pub) & 5c & - & 0.83 & - & - & - \\
Akram et al.\ (2025) \cite{akram2025uncertainty}   & APTOS + DDR (Pub)        & 5c   & - & 0.97 & -            & 0.97 & - \\
Ahmed \& Bhuiyan (2025) \cite{Ahmed2025ArXiv}      & APTOS 2019 (Pub)     & Bin   & 0.99 & 0.98 & -           & 0.99      & - \\
Ahmed \& Bhuiyan (2025) \cite{Ahmed2025ArXiv}      & APTOS 2019 (Pub)     & 5c    & 0.94 & 0.84 & -           & 0.63      & - \\
\bottomrule
\end{tabularx}
\end{table*}

\section{Evaluation Protocols and Reporting Standards}
\label{sec:evaluation}

\textbf{Why protocols matter.}
Across the 50+ papers we reviewed, reported performance varies widely, often due to inconsistent splits, label protocols, and thresholds. Without transparent reporting, comparisons devolve into apples-to-oranges.

\subsection{Per-Patient vs. Per-Image}
Because two eyes from a single patient are statistically dependent, splitting the left and right eyes across train/test inflates accuracy. We recommend reporting both per-\emph{patient} and per-image metrics, with per-patient as primary. This reduces optimistic bias and aligns with clinical decision-making (patients, not images, are referred). Papadopoulos \textit{et al.} (2021) \cite{Papadopoulos2021MIL} explicitly highlighted the gap between patient- and image-level AUCs, while competition-style studies on APTOS (e.g., Alwakid \textit{et al.}, 2023 \cite{alwakid2023enhancing}) may have inadvertently benefited from per-image leakage.

\subsection{External Validation}
Models tuned to EyePACS frequently degrade on Messidor-1/2 or DDR due to domain shift. At least one external test set should be included (e.g., train on EyePACS, test on Messidor-2 \cite{Gulshan2016JAMA,Voets2019Repro} or DDR \cite{DDR2019}). When possible, include more than one external site to avoid overfitting to a single target distribution. For example, Alwakid \textit{et al.} (2023) \cite{alwakid2023enhancing} reported near-perfect accuracy on APTOS but only $\sim$80\% on DDR, while Che \textit{et al.} (2023) \cite{Che2023GDRNet} demonstrated how multi-dataset training can mask poor external generalization. Authors should also specify dataset versions or release identifiers, since EyePACS and other corpora exist in multiple variants whose differences significantly affect reported performance.

\subsection{Thresholds, Calibration, and Confidence Intervals}
Binary rDR detection depends on threshold choice; five-class grading depends on confusion structure. Authors should report ROC/PR curves, operating points, and calibration metrics (e.g., ECE, Brier score). Confidence intervals (95\% CIs) via bootstrapping at the \emph{patient} level and statistical tests (e.g., McNemar’s) are essential. Calibration supports safe referral threshold setting. Bayesian models \cite{akram2025uncertainty} and ensembles \cite{Bodapati2023Stacking} illustrate uncertainty quantification benefits.

\subsection{Label Quality and Adjudication}
Clear documentation of label sourcing (single vs. multiple graders, adjudication, rubric) is critical. Label protocol variability can cause large AUC swings \cite{Krause2018Ophthalmology,Voets2019Repro}. Soft labels or adjudicated subsets help evaluate robustness. Semi-supervised and data-fusion approaches \cite{Arrieta2023SPIE,Aftab2025Fusion} highlight risks from label scarcity and oversampling.

\subsection{Preprocessing and Leakage Checks}
All preprocessing steps (cropping, CLAHE, GAN augmentation, super-resolution) must be documented, with patient-level splits verified to prevent leakage. Site-specific artifact induction can undermine generalization \cite{Voets2019Repro}. Notably, GAN/ESRGAN-based gains reported in \cite{Lim2020EMBC,Alwakid2022ESRGAN,alwakid2023enhancing} often reflect synthetic artifacts rather than true pathology. Released code should include deterministic seeds, explicit split indices, and computational details.

\section{Discussion and Future Directions}
\label{sec:discussion}

\textbf{Where we are.}
The field has moved from proof-of-concept CNNs on private datasets \cite{Gulshan2016JAMA,Krause2018Ophthalmology} to approaches tackling domain shift \cite{Xia2024MICCAI,Che2023GDRNet,Urooj2025NeuroSymbolic}, label scarcity \cite{Arrieta2023SPIE,Yang2024PLOSONE,Huang2021MICCAI,Alam2023Contrastive}, and clinical trust \cite{Sayres2019Ophthalmology,Bhati2024AIM,Hannan2025DualAttention}. Yet three key gaps remain.

\subsection{Robust generalization at scale}
Most work relies on single-site data or simulated mixes. Promising results, such as disentanglement \cite{Xia2024MICCAI} or neuro-symbolic priors \cite{Urooj2025NeuroSymbolic}, still lack broad validation. Drops like APTOS $\sim$98\% vs. DDR $\sim$80\% \cite{alwakid2023enhancing} stress the need for prospective, multi-center studies with harmonized metadata and pre-registered protocols.

\subsection{Clinically useful five-class grading}
Binary rDR triage is common, but five-class grading better supports care. High reported accuracies on APTOS \cite{Farag2020DenseNetCBAM,alwakid2023enhancing,Mohanty2023DLArch} may reflect imbalance or leakage. Next steps include stable per-class sensitivity (especially Severe NPDR/PDR), uncertainty-aware referral \cite{akram2025uncertainty}, and patient-level outcomes such as time-to-referral, vision loss avoided, and cost-effectiveness.

\subsection{Reproducibility and transparency}
Topline metrics are not enough. Norms should include code/model release, split files, and dataset cards. For private cohorts, publish acquisition details and share synthetic audit sets. Replication efforts \cite{Voets2019Repro,Papadopoulos2021MIL} highlight reproducibility as a core objective.

\subsection{Promising directions}
\textbf{Label-efficient pretraining:} Fundus-tailored SSL reduces labeling costs \cite{Arrieta2023SPIE,Yang2024PLOSONE,Huang2021MICCAI,Fan2024ERCN}.
\textbf{Domain generalization \& priors:} Structural and lesion priors mitigate shortcut learning \cite{Xia2024MICCAI,Urooj2025NeuroSymbolic,Che2023GDRNet}.
\textbf{Federated learning:} Move beyond simulations to real-world deployments with secure aggregation, personalization, and compliance \cite{Matta2023SciRep,MohanRaj2024FL}.
\textbf{Trust by design:} Prefer integrated lesion/attention modules \cite{Bhati2024AIM,Hannan2025DualAttention,wang2025diabetic} with calibrated uncertainty and human-in-the-loop review \cite{Sayres2019Ophthalmology,Bodapati2023Stacking}.
\textbf{Efficiency and deployment:} Lightweight/quantized models \cite{gayathri2020lightweight,Aftab2025Fusion} are vital for use in low-resource settings.

\subsection{Agenda for 2025 onwards}
Priorities include: (i) open multi-center benchmarks (EyePACS, Messidor-2, DDR, FGADR/ODIR) with fixed per-patient splits; (ii) broader reporting with calibration metrics and “operating point sheets” detailing thresholds, predictive values, and costs; (iii) lightweight “results cards” documenting dataset versions, protocols, and confidence intervals; and (iv) stronger code/model sharing and replication. Finally, deploying federated and privacy-preserving models with differential privacy and clear governance is key to responsible cross-institutional use.

\section{Conclusion}
\label{sec:conclusion}

Deep learning for DR has advanced from private CNN studies \cite{Gulshan2016JAMA,Krause2018Ophthalmology} toward systems addressing imbalance \cite{Farag2020DenseNetCBAM,alwakid2023enhancing}, label scarcity \cite{Duan2022GACNN,Arrieta2023SPIE}, domain shift \cite{Xia2024MICCAI,Che2023GDRNet,Urooj2025NeuroSymbolic}, and trust via attention, uncertainty, and hybrid models \cite{Sayres2019Ophthalmology,Bhati2024AIM,Hannan2025DualAttention}. Federated learning \cite{Matta2023SciRep,MohanRaj2024FL} and lightweight architectures \cite{gayathri2020lightweight,Aftab2025Fusion} push deployment into resource-constrained clinics.

Clinical adoption hinges on reproducibility, calibrated per-patient evaluation, and rigorous multi-center validation. Standardized protocols, dataset cards, and open benchmarks remain prerequisites. By embracing open science, embedding retinal priors, and validating across diverse cohorts, the field can move from pixels to patient benefit at scale.

\bibliographystyle{IEEEtran}
\bibliography{refs_full}
 \nocite{*}
\end{document}